\pdfoutput=1

\documentclass[runningheads]{llncs}
\usepackage{array}

\usepackage{eccv}

\usepackage{eccvabbrv}

\usepackage{graphicx}
\usepackage{booktabs}
\usepackage{algorithm}
\usepackage{algorithmic}
\usepackage[accsupp]{axessibility}

\usepackage{hyperref}
\usepackage{orcidlink}
\usepackage{amsmath,amssymb,amsfonts,mathrsfs}
\usepackage{upgreek}
\usepackage{bbding}
\usepackage[dvipsnames,svgnames]{xcolor}
\usepackage{booktabs,tabularx,multirow}
\usepackage{graphicx}
\usepackage[labelfont=bf]{caption}
\usepackage{subcaption}
\usepackage{textgreek}
\usepackage{algorithm}
\usepackage{algorithmic}
\usepackage{setspace}
\usepackage{balance}
\usepackage{tikz}
\usepackage{ulem}
\usepackage{comment}
\usepackage{wrapfig}
\usepackage{array}
\usepackage[utf8]{inputenc}
\usepackage{bm}
\usepackage{wrapfig}
\usepackage[misc]{ifsym}

\graphicspath{{./}{figure/}}
\begin{document}

\title{DiscoVL: Unveiling \textit{Dis}entangled \textit{C}ross-Modal Representation Learning via \textit{O}rthogonal Adversarial Regularization for \textit{V}ision-\textit{L}anguage Models}

\titlerunning{DiscoVL}

\author{
Mengping Dong\inst{1}\textsuperscript{*}\orcidlink{0009-0000-7201-7985}
\and
Jinbao Li\inst{1}\textsuperscript{*}\orcidlink{0000-0002-2432-8807}
\and
Fei Li\inst{2}$^{\dagger}$\orcidlink{0000-0002-8440-359X}
}

\authorrunning{M.~Dong et al.}

\institute{Shandong Artificial Intelligence Institute, Qilu University of Technology (Shandong Academy of Sciences), Jinan, China \\
\email{\{dongmengping,lijb\}@qlu.edu.cn}\\
\and
University of Florida, Florida, USA\\
\email{leefly072@126.com}}

\maketitle
\renewcommand{\thefootnote}{}
\footnotetext{* Equal contribution. $^{\dagger}$ Corresponding author.}
\renewcommand{\thefootnote}{\arabic{footnote}}

\begin{abstract}
Pre-trained vision-language models excel across varied perception tasks, but adapting them to novel downstream settings without sacrificing generalization remains non-trivial.
Existing parameter-efficient prompt learning method often yields inconsistent representations and fails to account for semantic distribution shifts.
In this work, we present DiscoVL, a disentangled cross-modal representation learning framework that couples orthogonal adversarial regularization with structured cross-modal alignment for vision-language models.
To address the insufficient cross-modal interaction, our DiscoVL designs a multi-branch low-rank residual aligner that decomposes representations into subspaces and enables bidirectional cross-modal feedback between visual and textual streams at each layer.
Furthermore, while conventional triplet constraints overfit features to class centroids, we design an orthogonal regularization for adversarial triplet loss, which prevents centroid collapse and substantially boosts generalization.
Evaluations on 15 benchmarks demonstrate that DiscoVL delivers consistent improvements over state-of-the-art methods for base-to-novel generalization, cross-dataset evaluation, and few-shot learning.
\end{abstract}

\section{Introduction}
\label{sec:intro}
Vision-language models (VLMs) such as CLIP \cite{radford2021learning} learn joint representations from millions of image-text pairs and exhibit impressive transfer to tasks ranging from zero-shot classification \cite{hao2025task} to open-vocabulary detection \cite{du2022learning} and segmentation \cite{liang2023open}.
Prompt learning \cite{zhao2024learning} has emerged as the dominant parameter-efficient strategy, which freezes the backbone and learns a small set of tunable tokens, avoiding the cost and forgetting risk of full fine-tuning.

A fundamental but largely overlooked problem is that existing VLMs adaptation methods learn \textit{entangled} representations in which class-specific discriminative cues and class-agnostic transferable knowledge are tightly intertwined.
Representative approaches, including CoOp/CoCoOp \cite{zhou2022learning, zhou2022conditional} and PromptSRC \cite{khattak2023self}, regularize prompt learning using textual prototypes or handcrafted priors.
However, the resulting representations still entangle task-relevant patterns with broad semantic priors inherited from pre-training.
Recent multi-modal prompt learners, such as AAPL \cite{kim2024aapl}, HiCroPL \cite{zheng2025hierarchical}, and MMRL \cite{guo2025mmrl}, aim to stabilize training by injecting attributes, routing knowledge flows, or stacking multiple representation aligners, yet they still optimize a single entangled feature space without explicitly separating these two types of information.

\begin{wrapfigure}[20]{r}{0.5\textwidth}
    \centering
    \includegraphics[width=0.5\textwidth]{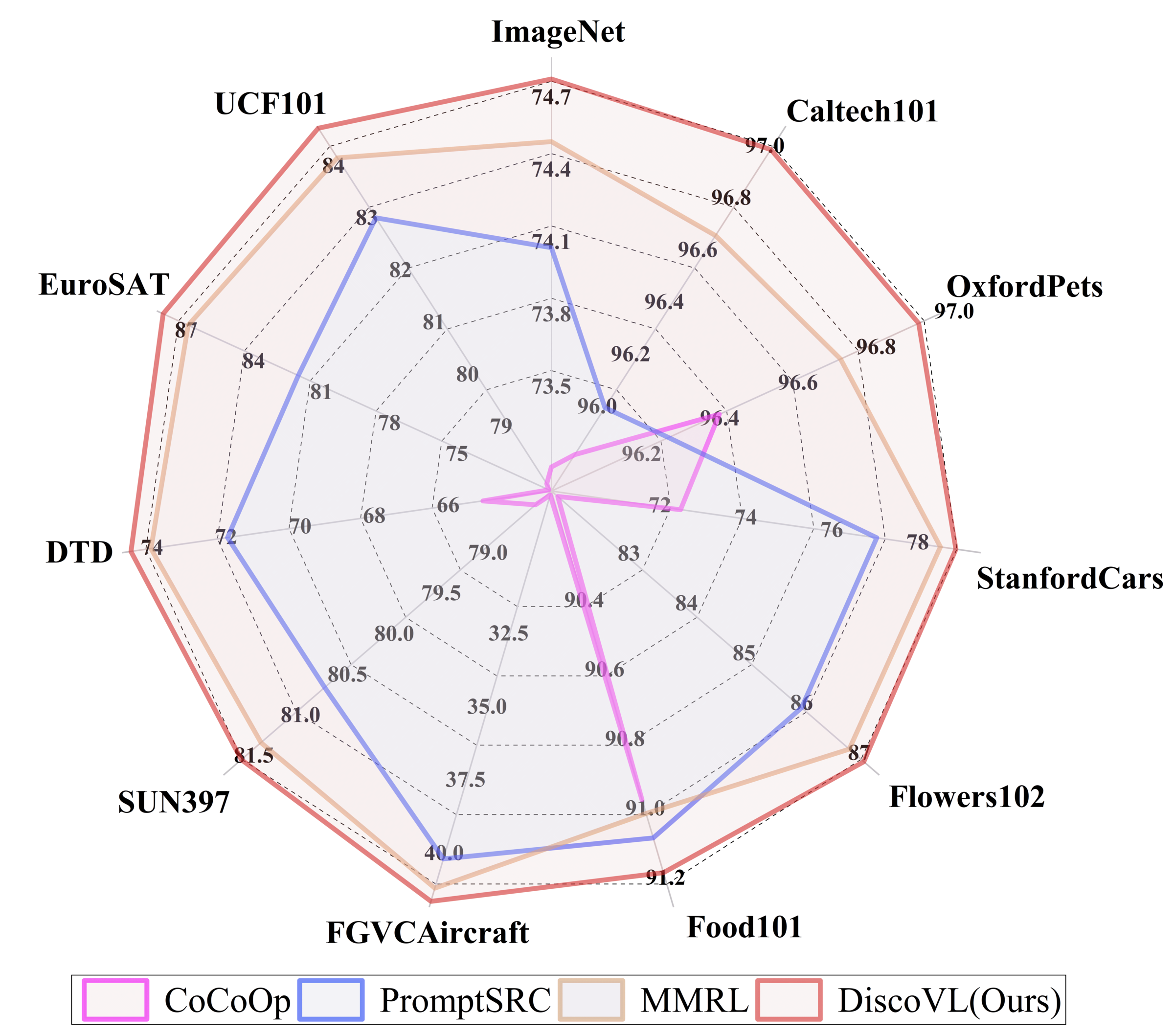}
    \caption{Comparison of the harmonic mean between previous state-of-the-art method and our DiscoVL across 11 diverse datasets for base-to-novel generalization.}
    \label{fig.1}
\end{wrapfigure}

This entanglement leads to three consequences:
1) class-specific adaptation contaminates the class-agnostic subspace, causing the representations to drift away from the pre-trained feature and eroding zero-shot transfer \cite{yao2023visual};
2) joint-space projections ignore modality-specific distribution gaps, producing inconsistent cross-modal representations that amplify this drift across layers;
and 3) representation tokens rapidly collapse onto class centroids, destroying the class-agnostic structure needed for novel-class generalization.

We address this root cause with \textit{DiscoVL}, a \textit{Dis}entangled \textbf{c}r\textbf{o}ss-modal representation learning framework for \textit{V}ision-\textit{L}anguage models adaptation.
The key idea is to explicitly factorize the adapted representation space into a \textit{shared, class-agnostic subspace} that preserves CLIP's transferable semantics and a \textit{task-specific subspace} that captures downstream discriminative cues.
This disentanglement is realized through two complementary modules.
First, we propose a \textit{Disentangled Cross-modal Representation Aligner (DCRA)}.
It employs a multi-branch residual architecture with subspace separation and a bidirectional cross-modal feedback loop.
This design progressively fuses textual and visual cues across layers while explicitly decoupling shared and task-specific information, thereby alleviating cross-modal and cross-layer inconsistencies.
Second, an \textit{Orthogonal Adversarial Representation Learning (OARL)} module enforces class-agnostic constraints in the representation space through orthogonal regularization and adversarial triplet objectives, preventing representation tokens from collapsing onto class centroids and preserving the pre-trained semantic representation.
Extensive experiments on 11 benchmarks, as shown in Fig. \ref{fig.1}, demonstrate that DiscoVL consistently surpasses prior prompt-learning methods on base-to-novel generalization.

We summarize our main contributions as follows:
\begin{itemize}
\item  We identify representation entanglement as a core bottleneck of existing VLMs prompt-learning methods, and propose disentangled cross-modal representation learning as a principled solution.
\item  A disentangled cross-modal representation aligner (DCRA) with multi-branch low-rank subspaces and bidirectional cross-modal feedback, which separates shared and task-specific representations while capturing complementary modality cues across layers.
\item  Orthogonal adversarial representation learning (OARL) that preserves class-agnostic knowledge through orthogonal triplet tokens and adversarial constraints, effectively suppressing token collapse.
\item  Comprehensive evaluations on 15 datasets showing consistent, significant gains over state-of-the-art prompt-learning baselines, with dedicated disentanglement analyses confirming the intended subspace separation.
\end{itemize}

\section{Related work}
\label{Related work}
\textbf{Vision-language models.}
Vision-language models (VLMs) have rapidly become the backbone of multi-modal perception, owing to their ability to contrastively align paired image-text samples in a shared latent space \cite{vasu2025fastvlm}.
CLIP \cite{radford2021learning} demonstrated that large-scale supervision, 400M image-text pairs, enables strong zero-shot generalization, and ALIGN \cite{jia2021scaling} further confirmed the scaling law with 1.8B weakly curated pairs.
Subsequent pipelines such as BLIP \cite{li2022boost} and LLaVA \cite{liu2023visual} inherit CLIP-like visual encoders while augmenting them with frozen LLM decoders to support open-ended reasoning.
However, when these models are adapted to downstream domains, class-specific and class-agnostic information becomes entangled in a single feature space, causing semantic drift and degraded cross-domain transfer.
DiscoVL addresses this by explicitly disentangling the two types of information through a cross-modal aligner and orthogonal adversarial regularization.

\textbf{Fine-tuning VLMs.}
Prompt engineering, first explored for LLM adaptation, has emerged as the dominant parameter-efficient strategy for tailoring VLMs.
Early CLIP deployments relied on carefully designed textual templates \cite{radford2021learning}, whereas later work learns prompt embeddings in the continuous token space \cite{hao2025task}.
KgCoOp \cite{yao2023visual} anchors soft prompts with a handcrafted prior to combat forgetting, while MaPLe \cite{khattak2023maple} injects branch-aware prompts into the dual vision-language towers to capture heterogeneous cues.
PromptSRC \cite{khattak2023self} introduces self-regulating constraints to stabilize prompt optimization, and CoPrompt \cite{roy2024consistency} enforces predictive consistency between frozen and trainable heads.
Beyond generic prompts, TCP \cite{yao2024tcp} derives class-aware representations via textual knowledge embeddings, TextRefiner \cite{xie2025textrefiner} augments prompts with cached visual descriptions, and SurPL \cite{liu2025surrogate} synthesizes target-aware text features through a lightweight surrogate generator.
These methods all operate within an entangled feature space where class-specific adaptation inevitably corrupts class-agnostic semantics.
DiscoVL differs fundamentally: it factorizes the representation space into disentangled shared and task-specific subspaces and enforces their separation with orthogonal adversarial regularization, addressing the root cause of cross-domain degradation rather than its symptoms.

\section{Methodology}
\label{Methodology}
\begin{figure*}[!t]
\centering
\includegraphics[scale=0.17]{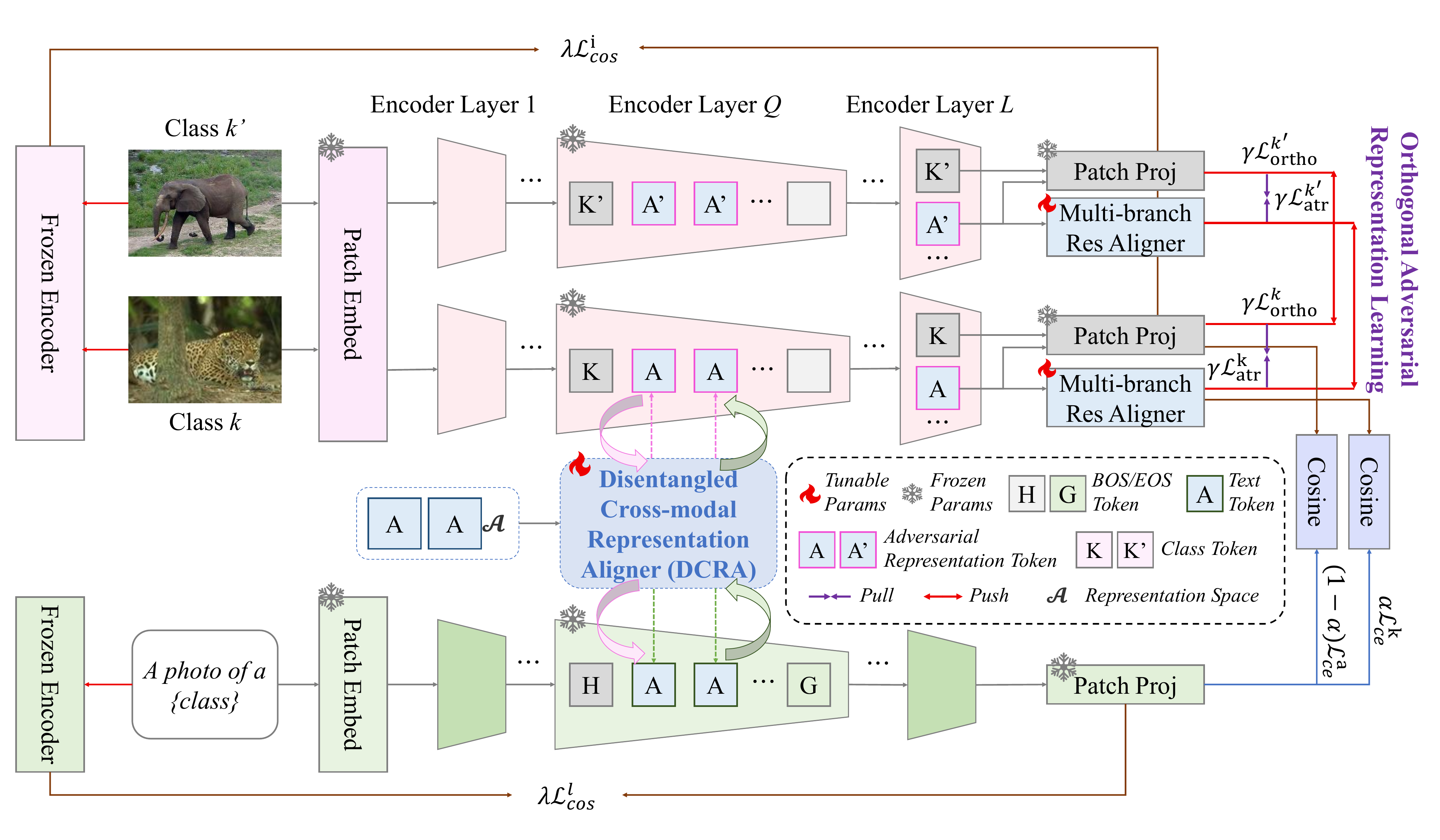}
\caption{Overview of DiscoVL, which disentangles representations into shared and task-specific subspaces. DCRA separates and aligns cross-modal cues via multi-branch residual structure and bidirectional feedback, while OARL enforces orthogonal adversarial constraints to prevent representation tokens from collapsing onto class centroids.}
\label{fig.2}
\end{figure*}

\subsection{Preliminaries}
\label{Preliminaries}
\textbf{Contrastive language-image pre-training.}
CLIP \cite{radford2021learning} comprises an image encoder $I$ and a text encoder $T$, each implemented as $L$ stacked transformer blocks $\{I_\ell\}_{\ell=1}^L$ and $\{T_\ell\}_{\ell=1}^L$.
Given an image $x \in \mathbb{R}^{H \times W \times 3}$, we partition it into $P$ patches and linearly project them to obtain $V_0 \in \mathbb{R}^{P \times d_i}$, where $d_i$ denotes the embedding dimension.
A learnable class token $k_0$ is prepended, and positional encoding is added before propagating the token sequence through the vision transformer $[k_\ell, V_\ell] = I_\ell([k_{\ell-1}, V_{\ell-1}]), \ell = 1, \dots, L$.
The final class token $k_L$ summarizes the entire image and is mapped into the shared vision-language space through the patch projection layer $Z_{img}$, yielding the image embedding $v = Z_{img}(k_L)$.
For the text branch, a prompt template is tokenized into $U$ tokens, which are embedded as $S_0 \in \mathbb{R}^{U \times d_t}$ with positional encoding, and $d_t$ is the embedding dimension.
Beginning-of-text and end-of-text tokens $(h_0, g_0)$ are appended, and the sequence is processed as
$[h_\ell, S_\ell, g_\ell] = T_\ell([h_{\ell-1}, S_{\ell-1}, g_{\ell-1}]), \ell = 1, \dots, L.$
The terminal token $g_L$ is projected with $Z_{txt}$ to produce $t\in \mathbb{R}^d$.
Given an image embedding $v\in \mathbb{R}^d$ and class-specific text embeddings $\{t_k\}_{k=1}^K$ for $K$ classes, CLIP predicts the class posterior via
\begin{align}
\scriptsize
P(y = k \mid v) = \frac{\exp(sim(v, t_k) / \tau)}{\sum_{i=1}^K \exp(sim(v, t_i) / \tau)}
\label{eq1}
\end{align}
where $sim(\cdot)$ denotes cosine similarity and $\tau$ is the temperature coefficient.
The DiscoVL training procedure is outlined in Algorithm \ref{alg:discovl}.

\begin{algorithm}[t]
\begin{minipage}{1.0\textwidth}
\algsetup{linenosize=\tiny}
\small
\caption{Training DiscoVL (DCRA + OARL)}
\label{alg:discovl}
\begin{algorithmic}[1]
\STATE \textbf{Input:} Pre-trained CLIP image encoder $I$, text encoder $T$, dataset $\mathcal{D}$, representation tokens $A$, injection layer $Q$
\STATE \textbf{Initialize:} DCRA parameters $\{F_{\text{shared}}^m, J_\ell^e, D_\ell^e, g(\cdot), \rho(\ell)\}$, projectors $Z_{\text{img}}^k$, $Z_{\text{img}}^a$, $Z_{T2I}$, $Z_{I2T}$
\FOR{each training iteration}
    \STATE Sample mini-batch $\{(x_i, y_i)\}$ from $\mathcal{D}$
    \STATE Encode text prompts with $T$ to obtain text feature $t$ and text tokens
    \STATE Encode image $x_i$ with $I$ and inject representation tokens $A$ from layer $Q$
    \STATE Apply DCRA to obtain image class token $k_L$, representation tokens $\{A_\ell^{i'}\}$
    \STATE Compute representation feature $a_L = \frac{1}{M}\sum_{j=1}^{M} A_{L,j}^{i'}$
    \STATE Project features $f_k = Z_{\text{img}}^k(k_L), f_a = Z_{\text{img}}^a(a_L)$
    \STATE Compute cross-modal logits via cosine similarity between $(f_k, t)$ and $(f_a, t)$
    \STATE Compute adversarial triplet loss $L_{\text{atr}}^{k}, L_{\text{atr}}^{k'}$
    \STATE Compute orthogonal regularization $L_{\text{ortho}}^{k}, L_{\text{ortho}}^{k'}$
    \STATE Compute classification losses $L_{\text{ce}}^{k}, L_{\text{ce}}^{a}$
    \STATE Compute cosine consistency with frozen CLIP $L_{\text{cos}}^{i}, L_{\text{cos}}^{l}$
    \STATE Update trainable parameters by minimizing $L_{\text{total}}$
\ENDFOR
\end{algorithmic}
\end{minipage}
\end{algorithm}

\textbf{Multi-modal representation learning.} MMRL \cite{guo2025mmrl} augments CLIP with a learnable representation space $\mathcal{A}$ that injects shared tokens into upper transformer layers of encoders.
Given representation tokens $A \in \mathbb{R}^{M \times d_a}$, where $M$ is the token number and $d_a$ is the representation space dimension.
A lightweight projector $\mathcal{G}(\cdot)$ maps them into modality-specific latent spaces, enabling cross-modal interaction without altering the frozen lower layers:
\begin{align}
\scriptsize
A^i = \{ A^i_\ell \}_{\ell=Q-1}^{L-1}, A^i_\ell = \mathcal{G}^i_\ell(A),
A^l = \{ A^l_\ell \}_{\ell=Q-1}^{L-1}, A^l_\ell = \mathcal{G}^l_\ell(A)
\label{eq2}
\end{align}
where $A^i_\ell \in \mathbb{R}^{M \times d_i}$ and $A^l_\ell \in \mathbb{R}^{M \times d_t}$ denote the injected tokens for the image and text branches at layer $\ell+1$, and $Q$ marks the first layer that receives the additional context.
Restricting the injections to higher layers preserves low-level priors learned during contrastive pre-training while allowing high-level semantics to adapt.
For image encoder $I$:
\begin{align}
\scriptsize
[k_\ell, V_\ell] = I_\ell(k_{\ell-1}, V_{\ell-1}), \quad \ell = 1, \dots, Q-1 \\
[k_\ell,\, \_,\, V_\ell] = I_\ell([k_{\ell-1}, A^i_{\ell-1}, V_{\ell-1}]), \quad \ell = Q, \dots, L-1 \\
[k_\ell, A^i_\ell, V_\ell] = I_\ell([k_{\ell-1}, A^i_{\ell-1}, V_{\ell-1}]), \quad \ell = L
\label{eq5}
\end{align}

For text encoder $T$, $A^l_\ell$ is prepended to the original sequence.
\begin{align}
\scriptsize
[h_\ell, S_\ell, g_\ell] = T_\ell([h_{\ell-1}, S_{\ell-1}, g_{\ell-1}]), \quad \ell = 1, \dots, Q-1 \\
[h_\ell, \_,\, S_\ell, g_\ell] = T_\ell([h_{\ell-1}, A^l_{\ell-1}, S_{\ell-1}, g_{\ell-1}]), \quad \ell = Q, \dots, L-1 \\
[h_\ell, A^l_\ell, S_\ell, g_\ell] = T_\ell([h_{\ell-1}, A^l_{\ell-1}, S_{\ell-1}, g_{\ell-1}]), \quad \ell = L
\label{eq8}
\end{align}

\subsection{Disentangled cross-modal representation aligner}
\begin{figure}[ht]
    \centering
    \begin{subfigure}[t]{0.44\textwidth}
        \centering
        \includegraphics[width=\linewidth]{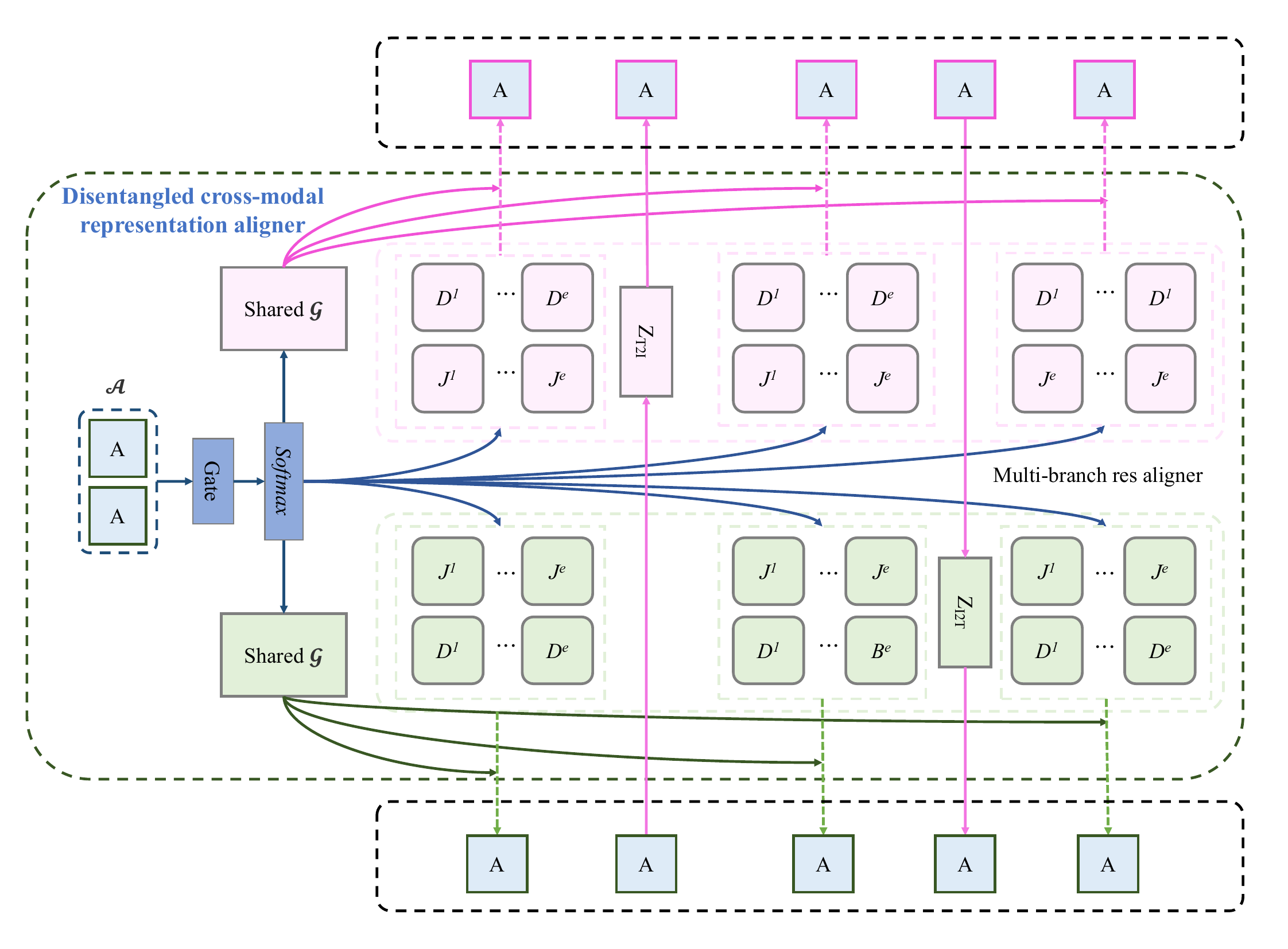}
        \caption{}
        \label{fig:sub6}
    \end{subfigure}
    \hfill
    \begin{subfigure}[t]{0.48\textwidth}
        \centering
        \includegraphics[width=\linewidth]{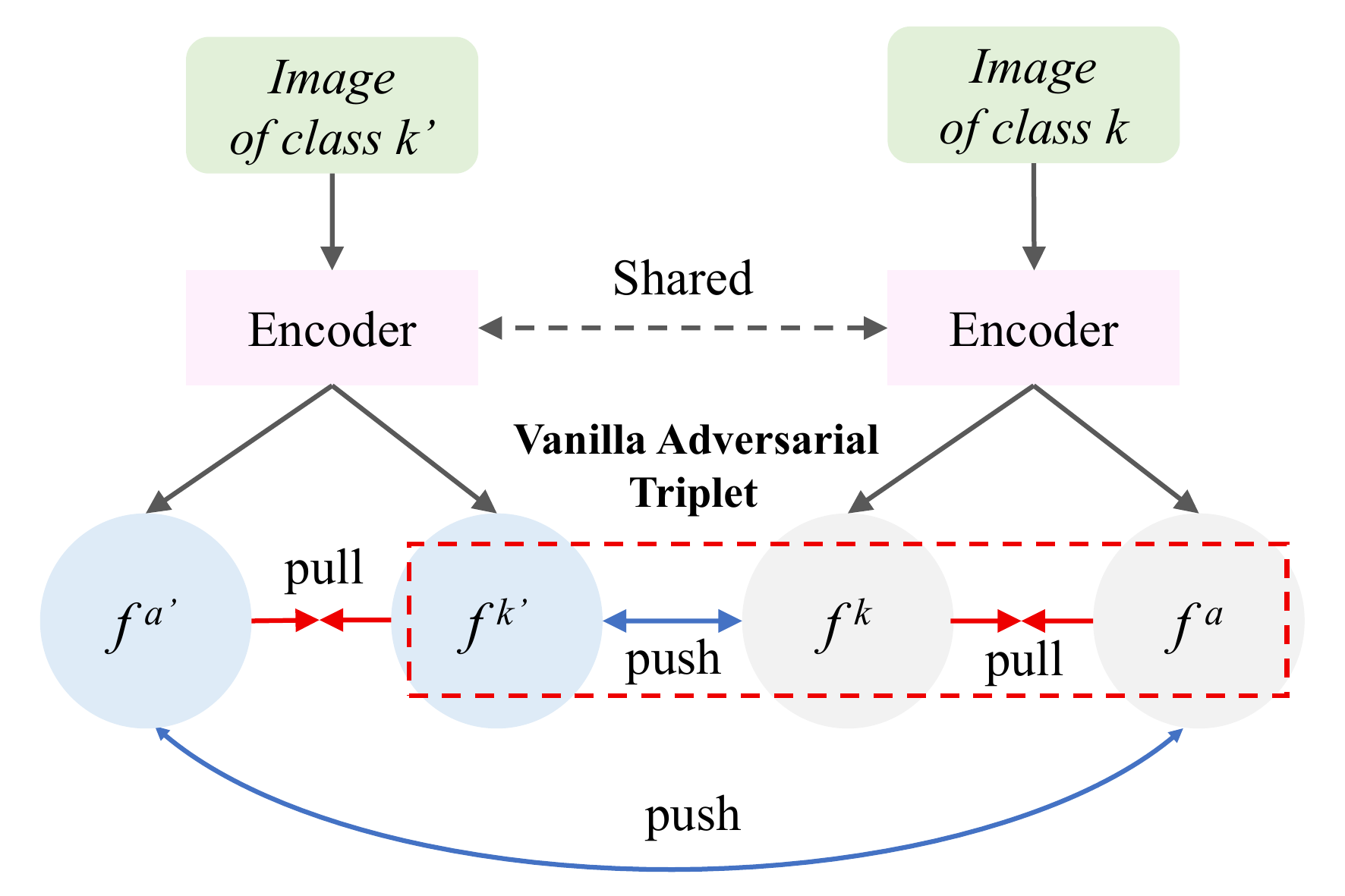}
        \caption{}
        \label{fig:sub3}
    \end{subfigure}
    \caption{Main component. (a) Disentangled cross-modal representation aligner, which contains a multi-branch residual aligner and bi-directional feedback between different image-text modalities. (b) Adversarial triplet regularization. The red dashed box is vanilla adversarial triplet regularization.}
    \label{fig:combined}
\end{figure}

Fig. \ref{fig.2} summarizes the overview of DiscoVL.
The images are tokenized into a sequence of patches, which are processed through a frozen CLIP vision backbone.
Starting from layer $Q$, class prompts are encoded by the text encoder, and both modalities meet in a lightweight projector $\mathcal{G}(\cdot)$ that lives entirely in the representation space.
Cosine similarities between projected tokens deliver the logits that are optimized jointly with the orthogonal adversarial learning objective.
Concretely, MMRL++ \cite{guo2026mmrl++} factorizes each projector into (i) a globally shared kernel that captures layer-invariant knowledge and (ii) a lightweight residual that absorbs layer-specific nuances.
For modality $m \in \{i,l\}$ and layers $\ell \in \{Q-1, \dots, L-1\}$, the factorization of projection weights is expressed as:
\begin{align}
\scriptsize
F_\ell^m = F_{\text{shared}}^m + J_\ell D_\ell, J_\ell \in \mathbb{R}^{d_a \times a_1}, D_\ell \in \mathbb{R}^{a_1 \times d_m}
\label{eq9}
\end{align}
where $a_1$ denotes the rank of the low-rank factorization.
$F_{\text{shared}}^m \in \mathbb{R}^{d_a \times d_m}$ is the globally shared component across layers, while $J_\ell D_\ell$ captures the layer-specific residual knowledge.
To better focus on the most relevant subspaces for each layer and modality, we extend this approach to a multi-branch residual aligner, as depicted in Fig. \ref{fig:sub6}.
This extension allows each layer to dynamically select different subspaces depending on the input representation, improving the model's capacity to capture diverse aspects of the data. The residual component is now represented as a weighted sum over $E$ subspaces, where each subspace is associated with a distinct low-rank factorization:
\begin{align}
\scriptsize
\Delta F_\ell^m = \sum_{e=1}^{E} o_\ell^e \, (J_\ell^e D_\ell^e), J_\ell^e \in \mathbb{R}^{d_a \times a_1}, D_\ell^e \in \mathbb{R}^{a_1 \times d_m}
\label{eq11}
\end{align}
where $J_\ell^e, D_\ell^e$ are branch-specific low-rank components and $o_\ell^e$ are attention weights that are dynamically predicted by a gating mechanism.
\begin{align}
\scriptsize
o_\ell = \text{\textit{Softmax}}(g(F_\ell^m))
\label{eq12}
\end{align}
where $g(\cdot)$ is a learned function that outputs the attention weights based on the current representation $F_\ell^m$.
This gating mechanism allows the model to adaptively emphasize the most relevant subspace for each layer by assigning higher attention to subspaces that are most aligned with the current data.
The designing of multiple branches enables the model to focus on distinct subspaces for each layer, allowing it to capture more complex and diverse relationships within and between modalities, while maintaining a compact and efficient parameterization.
For the patch projection layer of representation, we define its weights as:
\begin{align}
\scriptsize
F_a = F_i^k + J_a D_a, J_a \in \mathbb{R}^{d_i \times a_2}, D_a \in \mathbb{R}^{a_2 \times d}
\label{eq13}
\end{align}

To further tighten alignment, DCRA injects bi-directional feedback between different modalities.
In the lower half of the stack we enrich visual tokens with textual cues through a projection $Z_{T2I}$, controlled by a learned fusion gate $\rho(\ell)$:
\begin{align}
\scriptsize
 A_\ell^{i^{'}} = (1-\rho(\ell))F_\ell^i(A_\ell^i)+\rho(\ell)Z_{T2I}(F_\ell^l(A_\ell^l))
\label{eq14}
\end{align}

In the upper half, we reverse the process by sending visual summaries back to the text stream through $Z_{I2T}$:
\begin{align}
\scriptsize
 A_\ell^{l^{'}} = (1-\rho(\ell))F_\ell^l(A_\ell^l)+\rho(\ell)Z_{I2T}(F_\ell^i(A_\ell^{i^{'}}))
\label{eq15}
\end{align}

The fusion gate $\rho(\ell)$ is a learned function that enables the model to determine, at each layer $\ell$, the balance between preserving the original modality and fusing it with the other modality.
This adaptive mechanism allows DiscoVL to focus on the most discriminative representation while prioritizing the modality that provides the most relevant information at each depth.

\subsection{Orthogonal adversarial representation learning}
Adversarial triplet learning \cite{kim2024aapl} remains a strong mechanism for separating class semantics, but in practice we observe that representation tokens quickly collapse toward class centroids, which limits their ability to generalize to new categories.
To solve the issue, we design an \textit{adversarial representation triplet loss} that explicitly regulates cross-class interactions between image features and representation features.
Formally, let $f^k$ denote the image feature of class $k$ obtained from the image encoder, and $f^a$ its corresponding adversarial representation from the unified embedding space.
As depicted in Fig. \ref{fig:sub3}, for each anchor $f^k$ belonging to class $k$, we construct positive/negative pairs as: (1) Positive: the aligned representation feature $f^a$ from the same class $k$; (2) Negative: an image feature $f^{k'}$ sampled from a distinct class $k' \neq k$.
The loss is defined as follows, where $m$ is a margin parameter.
\begin{align}
\scriptsize
L_{\text{atr}}^k =
\max \big(0, \; \| f^k - f^a \|_2^2
- \| f^k - f^{k'} \|_2^2 + m \big), \\
L_{\text{atr}}^{k'} =
\max \big(0, \; \| f^{k'} - f^{a'} \|_2^2
- \| f^{k'}- f^k \|_2^2 + m \big)
\label{eq16}
\end{align}

The adversarial constraint aligns each image with its representation to ensure semantic consistency while separating features of different classes to prevent class-specific entanglement.
However, redundant features can still be exploited by adversarial perturbations, which manipulate multiple correlated features simultaneously and cause large output deviations.
Orthogonal regularization promotes feature independence, reducing redundancy and ensuring that each feature captures distinct information.
\begin{align}
\scriptsize
L_{\text{ortho}}^{k} = \zeta \cdot\left\|I-\frac{X^{k}(X^{k})^T}{\|X^{k}\|^2}\right\|_F^2
\label{eq17}
\end{align}
where $\zeta$ is the regularization hyperparameter.
$X^{k} \in \mathbb{R}^{M \times d_i}$ is the matrix representing the $f_k$.
$I$ is the identity matrix.
$\| \cdot \|_F $ is the Frobenius norm, which represents the square root of the sum of squares of all elements in the matrix.

In our framework, representation tokens are designed to capture dataset-specific adaptations, while the class token preserves the pre-trained knowledge from CLIP.
During training, we optimize both the representation tokens and the original class token, but prioritize the representation tokens to adapt them to the target dataset while retaining pre-trained knowledge.
Specifically, for the image encoder $I$, processing an input through $L$ transformer layers yields a class token $k_L \in \mathbb{R}^{d_i}$ and a set of representation tokens $ A_\ell^{i^{'}} \in \mathbb{R}^{M \times d_i}$.
The final representation feature $a_L$ is computed by averaging across the $M$ tokens.
Both class and representation tokens are projected into the shared vision-language latent space via patch projection layers:
\begin{align}
\scriptsize
f_k = Z_{img}^k(k_L), \quad f_a = Z_{img}^a(a_L)
\label{eq18}
\end{align}
where $Z_{img}^k$ is the frozen CLIP patch projection for class features, and $Z_{img}^a$ is a trainable projection for representation features.
For the text encoder $T$, we map the end-of-text token $g_L$ into the shared vision-language space $t=Z_{txt}(g_L)$ and compute cross-entropy losses:
\begin{align}
\scriptsize
L^k_{\text{ce}} = - \sum_{k=1}^{K} y_k \log p(y=k \mid f_k), \quad
L^a_{\text{ce}} = - \sum_{k=1}^{K} y_k \log p(y=k \mid f_a)
\label{eq19}
\end{align}
where $y_k$ is an indicator variable taking the value 1 if the image belongs to class $k$, and 0 otherwise.
To further preserve the generalization ability of class features, we maximize the cosine similarity between the updated features $(f_k, t)$ and the frozen CLIP features $(v, t_0)$:
\begin{align}
\scriptsize
L^{i}_{\text{cos}} = 1 - \frac{f_k \cdot v}{\|f_k\|\|v\|}, \quad
L^{l}_{\text{cos}} = 1 - \frac{1}{K} \sum_{k=1}^{K} \frac{t_k \cdot t_0^k}{|t_k||t_0^k|}
\label{eq20}
\end{align}

The total loss function are as follows:
\begin{align}
\scriptsize
L_{\text{total}} =
\alpha L^k_{\text{ce}} + (1-\alpha) L^a_{\text{ce}} +
\beta (L^i_{\text{cos}} + L^l_{\text{cos}}) +
\gamma (L_{\text{atr}}^{k} + L_{\text{atr}}^{k'} + L_{\text{ortho}}^{k} + L_{\text{ortho}}^{k'})
\label{eq21}
\end{align}
where $\alpha$, $\beta$, and $\gamma$ are hyper-parameters controlling the relative weights of classification, generalization, and orthogonal adversarial triplet regularization.
We combine the class features and the task-specific representation features to compute the probability for the base class \cite{guo2025mmrl}:
\begin{align}
\scriptsize
p(y=k \mid x) = \alpha \cdot p(y=k \mid f_k) + (1-\alpha) \cdot p(y=k \mid f_a)
\label{eq22}
\end{align}

For novel classes, only the class token is used for prediction, as it retains the general knowledge of the pre-trained CLIP.
Adversarial triplet learning representation tokens adapt to the dataset and are robust across classes, while the class token maintains generalization for new classes.
\begin{align}
\scriptsize
p(y=k \mid x) = p(y=k \mid f_k)
\label{eq23}
\end{align}

\section{Experiments}
\label{Experiments}
\subsection{Experimental setup}
\label{Experimental Setup}
\textbf{Evaluation and datasets.}
We follow the previous work \cite{guo2025mmrl,khattak2023self} to thoroughly evaluate DiscoVL across four tasks with several state-of-the-art methods: base-to-novel generalization, cross-dataset evaluation, few-shot learning, and domain generalization.
For those benchmarks, we use the same datasets as followed by previous works \cite{khattak2023self,guo2025mmrl}.
For base-to-novel generalization, cross-dataset evaluation, and few-shot learning, we conduct experiments on 11 datasets from diverse domains: generic object classification on ImageNet \cite{deng2009imagenet} and Caltech101 \cite{fei2004learning}; fine-grained classification on Oxford-Pets \cite{parkhi2012cats}, StanfordCars \cite{krause20133d}, Flowers102 \cite{nilsback2008automated}, Food101 \cite{bossard2014food}, and FGVCAircraft \cite{maji2013fine}; scene recognition on SUN397 \cite{xiao2010sun}; action recognition on UCF101 \cite{soomro2012ucf101}; texture classification on DTD \cite{cimpoi2014describing}; and satellite imagery recognition on EuroSAT \cite{helber2019eurosat}.
We evaluate DiscoVL's performance in domain generalization by testing it on four variants of ImageNet: ImageNet-V2 \cite{recht2019imagenet}, ImageNet-Sketch \cite{wang2019learning}, ImageNet-A \cite{hendrycks2021natural}, and ImageNet-R \cite{hendrycks2021many}.
Although these datasets share the same class structure as ImageNet, they exhibit significant differences in data distribution.

\begin{table*}[!t]
\caption{Base-to-novel generalization results. HM indicates the harmonic mean. The best and second best results are marked in bold and underline. DiscoVL consistently improves base class performance while preserving generalization to novel classes.}
\label{tab:base2}
\centering
\begin{minipage}{.24\linewidth}
\resizebox{\textwidth}{!}{
    \begin{tabular}{c|c c|c} \toprule
    \centering
    Method & Base & Novel & \bf{HM}\\ \midrule
    CoOp \cite{zhou2022learning}                  & 82.69 & 63.22 & 71.66 \\
    CoCoOp \cite{zhou2022conditional}             & 80.47 & 71.69 & 75.83 \\
    KgCoOp \cite{yao2023visual}                   & 80.73 & 73.60 & 77.00 \\
    MaPLe \cite{khattak2023maple}                 & 82.28 & 75.14 & 78.55 \\
    PromptSRC \cite{khattak2023self}              & 84.26 & 76.10 & 79.97 \\
    TCP \cite{yao2024tcp}                         & 84.13 & 75.36 & 79.51 \\
    MMA \cite{yang2024mma}                        & 83.20 & 76.80 & 79.87 \\
    AAPL \cite{kim2024aapl}                       & 80.27 & 72.17 & 76.01 \\
    CoPrompt \cite{roy2024consistency}            & 84.00 & 77.23 & 80.48 \\
    TextRefiner \cite{xie2025textrefiner}         & 79.74 & 74.32 & 76.94  \\
    RAda \cite{chen2025one}                       & 84.32 & 76.25 & 80.08 \\
    2SFS \cite{farina2025rethinking}              & 85.55 & 75.48 & 80.20  \\
    SkipT. \cite{wu2025skip}                      & 85.04 & \underline{77.53} & 81.11 \\
    MMRL  \cite{guo2025mmrl}                      & \underline{85.68} & 77.16 & \underline{81.20} \\
    \midrule
    \textbf{DiscoVL (Ours)}                         & \textbf{85.98} & \textbf{77.61} & \textbf{81.58
    }\\
    $\Delta$                                      & +0.30 & +0.45 & +0.38 \\
    \bottomrule
    \end{tabular}
    }
    \subcaption{Average}
\end{minipage}
\begin{minipage}{.24\linewidth}
\centering
    \resizebox{\textwidth}{!}{
    \begin{tabular}{c|c c|c}\toprule
    \centering
    Method & Base & Novel & \bf{HM}\\\midrule
    CoOp \cite{zhou2022learning}                & 98.00 & 89.81 & 93.73 \\
    CoCoOp \cite{zhou2022conditional}           & 97.96 & 93.81 & 95.84 \\
    KgCoOp \cite{yao2023visual}                 & 97.72 & 94.39 & 96.03  \\
    MaPLe \cite{khattak2023maple}               & 97.74 & 94.36 & 96.02 \\
    PromptSRC \cite{khattak2023self}            & 98.10 & 94.03 & 96.02 \\
    TCP \cite{yao2024tcp}                       & 98.23 & 94.67 & 96.42  \\
    MMA \cite{yang2024mma}                      & 98.40 & 94.00 & 96.15 \\
    AAPL \cite{kim2024aapl}                     & 97.87 & 95.10 & 96.46 \\
    CoPrompt \cite{roy2024consistency}          & 98.27 & 94.90 & 96.55 \\
    TextRefiner \cite{xie2025textrefiner}       & 98.13 & 94.43 & 96.24 \\
    RAda \cite{chen2025one}                     & 98.06 & 93.56 & 95.76 \\
    2SFS \cite{farina2025rethinking}            & 98.71 & 94.43 & 96.52 \\
    SkipT. \cite{wu2025skip}                    & 98.50 & \textbf{95.33} & \underline{96.89} \\
    MMRL  \cite{guo2025mmrl}                    & \underline{98.97} & 94.50 & 96.68 \\
    \midrule
    \textbf{DiscoVL (Ours)}                       & \textbf{99.13} & \underline{94.97} & \textbf{97.01}  \\
    $\Delta$                                    & +0.16 & +0.47 & +0.33 \\
    \bottomrule
    \end{tabular}
    }
    \subcaption{Caltech101}
\end{minipage}
\begin{minipage}{.24\linewidth}
    \resizebox{\textwidth}{!}{
    \begin{tabular}{c|c c|c}\toprule
    \centering
    Method & Base & Novel & \bf{HM}\\\midrule
    CoOp \cite{zhou2022learning}                 & 79.44 & 41.18 & 54.24 \\
    CoCoOp \cite{zhou2022conditional}            & 77.01 & 56.00 & 64.85 \\
    KgCoOp \cite{yao2023visual}                  & 77.55 & 54.99 & 64.35  \\
    MaPLe \cite{khattak2023maple}                & 80.36 & 59.18 & 68.16 \\
    PromptSRC \cite{khattak2023self}             & 83.37 & 62.97 & 71.75 \\
    TCP \cite{yao2024tcp}                        & 82.77 & 58.07 & 68.25 \\
    MMA \cite{yang2024mma}                       & 83.20 & 65.63 & 73.38 \\
    AAPL \cite{kim2024aapl}                      & 73.90 & 53.43 & 62.02 \\
    CoPrompt \cite{roy2024consistency}           & 83.13 & 64.73 & 72.79 \\
    TextRefiner \cite{xie2025textrefiner}        & 75.35 & 58.09 & 65.60  \\
    RAda \cite{chen2025one}                      & 82.06 & \underline{67.15} & 73.86 \\
    2SFS \cite{farina2025rethinking}             & 84.60 & 65.01 & 73.52  \\
    SkipT. \cite{wu2025skip}                     & 83.77 & \textbf{67.23} & \textbf{74.59} \\
    MMRL  \cite{guo2025mmrl}                     & \underline{85.67} & 65.00 & 73.82 \\
    \midrule
    \textbf{DiscoVL (Ours)}                        & \textbf{85.77} & 65.63 & \underline{74.36} \\
    $\Delta$                                     & +0.10 & +0.63 & +0.54 \\
    \bottomrule
    \end{tabular}
    }
    \subcaption{DTD}
\end{minipage}
\begin{minipage}{.24\linewidth}
\resizebox{\textwidth}{!}{
    \begin{tabular}{c|c c|c}\toprule
    \centering
    Method & Base & Novel & \bf{HM}\\\midrule
    CoOp \cite{zhou2022learning}                 & 92.19 & 54.74 & 68.69 \\
    CoCoOp \cite{zhou2022conditional}            & 87.49 & 60.04 & 71.21 \\
    KgCoOp \cite{yao2023visual}                  & 85.64 & 64.34 & 73.48 \\
    MaPLe \cite{khattak2023maple}                & 94.07 & 73.23 & 82.35 \\
    PromptSRC \cite{khattak2023self}             & 92.90 & 73.90 & 82.32 \\
    TCP \cite{yao2024tcp}                        & 91.63 & 74.73 & 82.32 \\
    MMA \cite{yang2024mma}                       & 85.46 & \underline{82.34} & 83.87 \\
    AAPL \cite{kim2024aapl}                      & 87.00 & 66.30 & 75.25 \\
    CoPrompt \cite{roy2024consistency}           & 94.60 & 78.57 & 85.84 \\
    TextRefiner \cite{xie2025textrefiner}        & 74.57 & 72.82 & 73.68  \\
    RAda \cite{chen2025one}                      & 96.48 & 74.69 & 84.20 \\
    2SFS \cite{farina2025rethinking}             & \textbf{96.91} & 67.09 & 79.29 \\
    SkipT. \cite{wu2025skip}                     & 92.47 & \textbf{83.00} & \underline{87.48} \\
    MMRL  \cite{guo2025mmrl}                     & 95.60 & 80.17 & 87.21 \\
    \midrule
    \textbf{DiscoVL (Ours)}                        & \underline{96.80} & 81.20 & \textbf{88.32} \\
    $\Delta$                                     & +1.20 & +1.03 & +1.11 \\
    \bottomrule
    \end{tabular}
    }
    \subcaption{EuroSAT}
\end{minipage}
\begin{minipage}{.24\linewidth}
\resizebox{\textwidth}{!}{
    \begin{tabular}{c|c c|c}\toprule
    \centering
    Method & Base & Novel & \bf{HM}\\\midrule
    CoOp \cite{zhou2022learning}                   & 40.44 & 22.30 & 28.75 \\
    CoCoOp \cite{zhou2022conditional}              & 33.41 & 23.71 & 27.74 \\
    KgCoOp \cite{yao2023visual}                    & 36.21 & 33.55 & 34.83 \\
    MaPLe \cite{khattak2023maple}                  & 37.44 & 35.61 & 36.50 \\
    PromptSRC \cite{khattak2023self}               & 42.73 & 37.87 & 40.15 \\
    TCP \cite{yao2024tcp}                          & 41.97 & 34.43 & 37.83 \\
    MMA \cite{yang2024mma}                         & 40.57 & 36.33 & 38.33 \\
    AAPL \cite{kim2024aapl}                        & 34.07 & 24.17 & 28.28 \\
    CoPrompt \cite{roy2024consistency}             & 40.20 & \textbf{39.33} & 39.76 \\
    TextRefiner \cite{xie2025textrefiner}          & 35.35 & 35.87 & 35.61  \\
    RAda \cite{chen2025one}                        & 41.72 & \underline{38.09} & 39.82 \\
    2SFS \cite{farina2025rethinking}               & \textbf{47.48} & 35.51 & 40.63 \\
    SkipT. \cite{wu2025skip}                       & 45.37 & 37.13 & 40.84 \\
    MMRL  \cite{guo2025mmrl}                       & 46.30 & 37.03 & \underline{41.15} \\
    \midrule
    \textbf{DiscoVL (Ours)}                          & \underline{46.40} & 37.70 & \textbf{41.60}  \\
    $\Delta$                                       & +0.10 & +0.67 & +0.45 \\
    \bottomrule
    \end{tabular}
    }
    \subcaption{FGVCAircraft}
\end{minipage}
\begin{minipage}{.24\linewidth}
\resizebox{\textwidth}{!}{
    \begin{tabular}{c|c c|c}\toprule
    \centering
    Method & Base & Novel & \bf{HM}\\\midrule
    CoOp \cite{zhou2022learning}                    & 88.33 & 82.26 & 85.19 \\
    CoCoOp \cite{zhou2022conditional}               & 90.70 & 91.29 & 90.99 \\
    KgCoOp \cite{yao2023visual}                     & 90.50 & 91.70 & 91.09 \\
    MaPLe \cite{khattak2023maple}                   & 90.71 & \underline{92.05} & 91.38 \\
    PromptSRC \cite{khattak2023self}                & 90.67 & 91.53 & 91.10 \\
    TCP \cite{yao2024tcp}                           & 90.57 & 91.37 & 90.97 \\
    MMA \cite{yang2024mma}                          & 90.13 & 91.30 & 90.71 \\
    AAPL \cite{kim2024aapl}                         & 90.70 & 91.60 & 91.15 \\
    CoPrompt \cite{roy2024consistency}              & \underline{90.73} & \textbf{92.07} & \textbf{91.40} \\
    TextRefiner \cite{xie2025textrefiner}           & \textbf{90.88} & 91.43 & 91.15 \\
    RAda \cite{chen2025one}                         & 90.35 & 91.49 & 90.92 \\
    2SFS \cite{farina2025rethinking}                & 89.11 & 91.34 & 90.21 \\
    SkipT. \cite{wu2025skip}                        & 90.67 & 92.03 & \underline{91.34} \\
    MMRL  \cite{guo2025mmrl}                        & 90.57 & 91.50 & 91.03 \\
    \midrule
    \textbf{DiscoVL (Ours)}                           & \underline{90.73} & 91.67 & 91.20 \\
    $\Delta$                                        & +0.16 & +0.17 & +0.17 \\
    \bottomrule
    \end{tabular}
    }
    \subcaption{Food101}
\end{minipage}
\begin{minipage}{.24\linewidth}
\resizebox{\textwidth}{!}{
    \begin{tabular}{c|c c|c}\toprule
    \centering
    Method & Base & Novel & \bf{HM}\\\midrule
    CoOp \cite{zhou2022learning}                   & 76.47 & 67.88 & 71.92 \\
    CoCoOp \cite{zhou2022conditional}              & 75.98 & 70.43 & 73.10 \\
    KgCoOp \cite{yao2023visual}                    & 75.83 & 69.96 & 72.78 \\
    MaPLe \cite{khattak2023maple}                  & 76.66 & 70.54 & 73.47 \\
    PromptSRC \cite{khattak2023self}               & 77.60 & 70.73 & 74.01 \\
    TCP \cite{yao2024tcp}                          & 77.27 & 69.87 & 73.38  \\
    MMA \cite{yang2024mma}                         & 77.31 & 71.00 & 74.02 \\
    AAPL \cite{kim2024aapl}                        & 76.53 & 70.57 & 73.43 \\
    CoPrompt \cite{roy2024consistency}             & 77.67 & 71.27 & 74.33 \\
    TextRefiner \cite{xie2025textrefiner}          & 76.84 & 70.54 & 73.56  \\
    RAda \cite{chen2025one}                        & \underline{77.96} & 70.23 & 73.89 \\
    2SFS \cite{farina2025rethinking}               & 77.71 & 70.99 & 74.20 \\
    SkipT. \cite{wu2025skip}                       & 77.73 & 70.40 & 73.89 \\
    MMRL  \cite{guo2025mmrl}                       & 77.90 & \underline{71.30} & \underline{74.45} \\
    \midrule
    \textbf{DiscoVL (Ours)}                          & \textbf{78.07} & \textbf{71.63} & \textbf{74.71} \\
    $\Delta$                                       & +0.17 & +0.33 & +0.26 \\
    \bottomrule
    \end{tabular}
    }
    \subcaption{ImageNet}
\end{minipage}
\begin{minipage}{.24\linewidth}
\resizebox{\textwidth}{!}{
    \begin{tabular}{c|c c|c}\toprule
    \centering
    Method & Base & Novel & \bf{HM}\\\midrule
    CoOp \cite{zhou2022learning}                & 97.60 & 59.67 & 74.06 \\
    CoCoOp \cite{zhou2022conditional}           & 94.87 & 71.75 & 81.71 \\
    KgCoOp \cite{yao2023visual}                 & 95.00 & 74.73 & 83.65 \\
    MaPLe \cite{khattak2023maple}               & 95.92 & 72.46 & 82.56 \\
    PromptSRC \cite{khattak2023self}            & 98.07 & 76.50 & 85.95 \\
    TCP \cite{yao2024tcp}                       & 97.73 & 75.57 & 85.23 \\
    MMA \cite{yang2024mma}                      & 97.77 & 75.93 & 85.48 \\
    AAPL \cite{kim2024aapl}                     & 95.10 & 70.63 & 81.06 \\
    CoPrompt \cite{roy2024consistency}          & 97.27 & 76.60 & 85.71 \\
    TextRefiner \cite{xie2025textrefiner}       & 95.92 & 74.33 & 83.76  \\
    RAda \cite{chen2025one}                     & 97.74 & 75.32 & 85.08 \\
    2SFS \cite{farina2025rethinking}            & 98.29 & 76.17 & 85.83 \\
    SkipT. \cite{wu2025skip}                    & 98.57 & 75.80 & 85.70 \\
    MMRL  \cite{guo2025mmrl}                    & \textbf{98.97} & \underline{77.27} & \underline{86.78} \\
    \midrule
    \textbf{DiscoVL (Ours)}                       & \underline{98.90} & \textbf{77.70} & \textbf{87.03} \\
    $\Delta$                                    & -0.07 & +0.43 & +0.25 \\
    \bottomrule
    \end{tabular}
    }
    \subcaption{Flowers102}
\end{minipage}
\begin{minipage}{.24\linewidth}
\resizebox{\textwidth}{!}{
    \begin{tabular}{c|c c|c}\toprule
    \centering
    Method & Base & Novel & \bf{HM}\\\midrule
    CoOp \cite{zhou2022learning}                    & 93.67 & 95.29 & 94.47 \\
    CoCoOp \cite{zhou2022conditional}               & 95.20 & 97.69 & 96.43 \\
    KgCoOp \cite{yao2023visual}                     & 94.65 & 97.76 & 96.18 \\
    MaPLe \cite{khattak2023maple}                   & 95.43 & 97.76 & 96.58 \\
    PromptSRC \cite{khattak2023self}                & 95.33 & 97.30 & 96.30 \\
    TCP \cite{yao2024tcp}                           & 94.67 & 97.20 & 95.92  \\
    MMA \cite{yang2024mma}                          & 95.40 & \underline{98.07} & 96.72 \\
    AAPL \cite{kim2024aapl}                         & 95.63 & 97.40 & 96.51 \\
    CoPrompt \cite{roy2024consistency}              & 95.67 & \textbf{98.10} & \underline{96.87} \\
    TextRefiner \cite{xie2025textrefiner}           & 95.27 & 97.65 & 96.45  \\
    RAda \cite{chen2025one}                         & 95.43 & 97.99 & 96.69 \\
    2SFS \cite{farina2025rethinking}                & 95.32 & 97.82 & 96.55 \\
    SkipT. \cite{wu2025skip}                        & 95.70 & 97.87 & 96.77 \\
    MMRL  \cite{guo2025mmrl}                        & \underline{95.90} & 97.60 & 96.74 \\
    \midrule
    \textbf{DiscoVL (Ours)}                           & \textbf{96.20} & 97.70 & \textbf{96.94} \\
    $\Delta$                                        & +0.30 & +0.10 & +0.20 \\
    \bottomrule
    \end{tabular}
    }
    \subcaption{OxfordPets}
\end{minipage}
\begin{minipage}{.24\linewidth}
\resizebox{\textwidth}{!}{
    \begin{tabular}{c|c c|c}\toprule
    \centering
    Method & Base & Novel & \bf{HM}\\\midrule
    CoOp \cite{zhou2022learning}                       & 78.12 & 60.40 & 68.13 \\
    CoCoOp \cite{zhou2022conditional}                  & 70.49 & 73.59 & 72.01 \\
    KgCoOp \cite{yao2023visual}                        & 71.76 & 75.04 & 73.36 \\
    MaPLe \cite{khattak2023maple}                      & 72.94 & 74.00 & 73.47 \\
    PromptSRC \cite{khattak2023self}                   & 78.27 & 74.97 & 76.58 \\
    TCP \cite{yao2024tcp}                              & 80.80 & 74.13 & 77.32 \\
    MMA \cite{yang2024mma}                             & 78.50 & 73.10 & 75.70 \\
    AAPL \cite{kim2024aapl}                            & 70.33 & 73.50 & 71.88 \\
    CoPrompt \cite{roy2024consistency}                 & 76.97 & 74.40 & 75.66 \\
    TextRefiner \cite{xie2025textrefiner}              & 71.40 & 70.90 & 71.15  \\
    RAda \cite{chen2025one}                            & 79.36 & 73.16 & 76.13 \\
    2SFS \cite{farina2025rethinking}                   & \underline{82.50} & 74.80 & \textbf{78.46} \\
    SkipT. \cite{wu2025skip}                           & \textbf{82.93} & 72.50 & 77.37 \\
    MMRL  \cite{guo2025mmrl}                           & 81.30 & \underline{75.07} & 78.06 \\
    \midrule
\textbf{DiscoVL (Ours)}                                  & 81.80 & \textbf{75.30} & \underline{78.42} \\
    $\Delta$                                           & +0.50 & +0.23 & +0.36 \\
    \bottomrule
    \end{tabular}
    }
    \subcaption{StanfordCars}
\end{minipage}
\begin{minipage}{.24\linewidth}
    \centering
\resizebox{\textwidth}{!}{
    \begin{tabular}{c|c c|c}\toprule
    Method & Base & Novel & \bf{HM}\\\midrule
    CoOp \cite{zhou2022learning}                   & 80.60 & 65.89 & 72.51 \\
    CoCoOp \cite{zhou2022conditional}              & 79.74 & 76.86 & 78.27 \\
    KgCoOp \cite{yao2023visual}                    & 80.29 & 76.53 & 78.36 \\
    MaPLe \cite{khattak2023maple}                  & 80.82 & 78.70 & 79.75 \\
    PromptSRC \cite{khattak2023self}               & 82.67 & 78.47 & 80.52 \\
    TCP \cite{yao2024tcp}                          & 82.63 & 78.20 & 80.35 \\
    MMA \cite{yang2024mma}                         & 82.27 & 78.57 & 80.38 \\
    AAPL \cite{kim2024aapl}                        & 79.65 & 76.90 & 78.25  \\
    CoPrompt \cite{roy2024consistency}             & 82.63 & \textbf{80.03} & \underline{81.31} \\
    TextRefiner \cite{xie2025textrefiner}          & 80.96 & 76.49 & 78.66  \\
    RAda \cite{chen2025one}                        & 82.58 & 78.77 & 80.63 \\
    2SFS \cite{farina2025rethinking}               & 82.59 & 78.91 & 80.70 \\
    SkipT. \cite{wu2025skip}                       & 82.40 & 79.03 & 80.68 \\
    MMRL  \cite{guo2025mmrl}                       & \underline{83.20} & 79.30 & 81.20 \\
    \midrule
    \textbf{DiscoVL (Ours)}                          & \textbf{83.30} & \underline{79.60} & \textbf{81.41}  \\
    $\Delta$                                       & +0.10 & +0.30 & +0.21  \\
    \bottomrule
    \end{tabular}
    }
    \subcaption{SUN397}
\end{minipage}
\begin{minipage}{.24\linewidth}
    \centering
    \resizebox{\textwidth}{!}{
    \begin{tabular}{c|c c|c}\toprule
    Method & Base & Novel & \bf{HM}\\\midrule
    CoOp \cite{zhou2022learning}                   & 84.69 & 56.05 & 67.46  \\
    CoCoOp \cite{zhou2022conditional}              & 82.33 & 73.45 & 77.64  \\
    KgCoOp \cite{yao2023visual}                    & 82.89 & 76.67 & 79.65  \\
    MaPLe \cite{khattak2023maple}                  & 83.00 & 78.66 & 80.77  \\
    PromptSRC \cite{khattak2023self}               & 87.10 & 78.80 & 82.74  \\
    TCP \cite{yao2024tcp}                          & 87.13 & \underline{80.77} & 83.83  \\
    MMA \cite{yang2024mma}                         & 86.23 & 80.03 & 82.20  \\
    AAPL \cite{kim2024aapl}                        & 82.20 & 74.27 & 78.03  \\
    CoPrompt \cite{roy2024consistency}             & 86.90 & 79.57 & 83.07  \\
    TextRefiner \cite{xie2025textrefiner}          & 82.52 & 75.01 & 78.59  \\
    RAda \cite{chen2025one}                        & 85.78 & 78.26 & 81.85  \\
    2SFS \cite{farina2025rethinking}               & 87.85 & 78.19 & 82.74  \\
    SkipT. \cite{wu2025skip}                       & 87.30 & \textbf{82.47} & \textbf{84.81}  \\
    MMRL  \cite{guo2025mmrl}                       & \underline{88.10} & 80.07 & 83.89  \\
    \midrule
    \textbf{DiscoVL (Ours)}                          & \textbf{88.67} & 80.63 & \underline{84.46}  \\
    $\Delta$                                       & +0.57 & +0.56 & +0.57  \\
    \bottomrule
    \end{tabular}
    }
    \subcaption{UCF101}
\end{minipage}
\end{table*}

\textbf{Implementation details.}
We follow prior studies \cite{khattak2023self,guo2025mmrl} and adopt a 16-shot learning setting across all experiments, except for the few-shot learning tasks.
The ViT-B/16 variant of the CLIP model serves as the visual backbone for all experimental setups \cite{guo2025mmrl}.
For each test image, we use the hand-crafted text prompt "a photo of a [class]."
Optimization is performed using the AdamW optimizer with an initial learning rate of 0.001.
Training on ImageNet for the base-to-novel generalization task spans 5 epochs, whereas training on the remaining datasets is conducted over 10 epochs.
For cross-dataset evaluation and domain generalization tasks, we perform training for a single epoch on ImageNet.
We report the base-class accuracy (\%), denoted as Base, the novel-class accuracy (\%), denoted as Novel, and their harmonic mean (\%), denoted as HM, to compare the effectiveness of different methods.
In the few-shot learning tasks, training is carried out for 5 epochs on ImageNet and 50 epochs for other datasets for 1, 2, 4, 8, 16 shot.
We test the performance of models after training, the average accuracy is reported over three independent runs.
We introduce $M = 5$ representation tokens starting from the Q-th Transformer layer, with $Q = 6$.
The low rank dimensions $a_1$ and $a_2$ are set to 4 and 64, respectively.
The parameter $\alpha$ is fixed at 0.7, and the parameter of $\beta$ is set to 0.5, the parameter $\gamma$ is 0.3.

\subsection{Base-to-novel generalization}
As reported in Table \ref{tab:base2}, DiscoVL achieves competitiveness across all key metrics, with improvements of 0.30\%, 0.45\%, and 0.38\% on base, novel, and HM, respectively, over the previous best model, MMRL.
DiscoVL provides a greater HM than all competitors in 7 out of 11 benchmarks and outperforms all competitors on average across datasets.
Notably, on EuroSAT, DiscoVL improves HM by 1.11\%.
Although it underperforms Skip Tuning by 1.80\% in novel-class accuracy, it achieves a 4.33\% gain in base-class accuracy, leading to a higher overall HM and a more balanced base–novel trade-off.

\begin{table*}[!t]
\caption{Comparisons with state-of-the-art methods on cross-dataset evaluation. Bold values indicate the best results. Overall, DiscoVL provides the highest average accuracy, indicating better generalization.}
\centering
\scalebox{0.62}{%
\begin{tabular}{@{}c|cccccccccccc@{}}
\toprule
\multirow{2}{*}{\textbf{Method}} & \textbf{Source} & \multicolumn{11}{c}{\textbf{Target dataset}} \\
\cmidrule(r){2-13}
& ImageNet& Caltech101 & Pets & Cars & Flowers102 & Food101 & Aircraft & SUN397 & DTD & EuroSAT & UCF101 & Average \\ \cmidrule(r){1-13}
CoOp \cite{zhou2022learning}                & 71.51 & 93.70 & 89.14 & 64.51 & 68.71 & 85.30 & 18.47 & 64.15 & 41.90 & 46.39 & 66.55 & 63.88  \\
CoCoOp \cite{zhou2022conditional}           & 71.02 & 94.43 & 90.14 & 65.32 & 71.88 & 86.06 & 22.94 & 67.36 & 45.73 & 45.37 & 68.21 & 65.74  \\
MaPLe \cite{khattak2023maple}               & 70.72 & 93.53 & 90.49 & 65.57 & 72.23 & 86.20 & 24.74 & 67.01 & 46.49 & 48.06 & 68.69 & 66.30  \\
PromptSRC \cite{khattak2023self}            & 71.27 & 93.60 & 90.25 & 65.70 & 70.25 & 86.15 & 23.90 & 67.10 & 46.87 & 45.50 & 68.75 & 65.81  \\
TCP \cite{yao2024tcp}                       & 71.40 & 93.97 & 91.25 & 64.69 & 71.21 & \underline{86.69} & 23.45 & 67.15 & 44.35 & 51.45 & 68.73 & 66.29  \\
MMA \cite{yang2024mma}                      & 71.00 & 93.80 & 90.30 & \underline{66.13} & 72.07 & 86.12 & 25.33 & \textbf{68.17} & 46.57 & 49.24 & 68.32 & 66.61  \\
AAPL \cite{kim2024aapl}                     & 71.37 & 94.17 & 90.73 & 65.10 & 71.67 & 86.00 & 23.03 & 66.80 & 44.80 & 41.83 & \underline{69.30} & 65.34  \\
CoPrompt \cite{roy2024consistency}          & 70.80 & 94.50 & 90.73 & 65.67 & 72.30 & 86.43 & 24.00 & 67.57 & \underline{47.07} & 51.90 & \textbf{69.73} & 67.00  \\
SkipT. \cite{wu2025skip}                    & \textbf{72.77} & 93.43 & 90.10 & 65.37 & 71.97 & 86.17 & 25.13 & 67.33 & \textbf{48.00} & \textbf{54.27} & 68.23 & 67.00  \\
MMRL \cite{guo2025mmrl}                     & 72.03 & \underline{94.67} & \underline{91.43} & 66.10 & \underline{72.77} & 86.40 & \underline{26.30} & 67.57 & 45.90 & \underline{53.10} & 68.27 & \underline{67.25}  \\
\midrule
\textbf{DiscoVL (Ours)}                       & \underline{72.10} & \textbf{94.73} & \textbf{91.90} & \textbf{66.30} & \textbf{72.87} & \textbf{86.70} & \textbf{26.47} & \underline{67.80} & 46.13 & 53.07 & 68.80 & \textbf{67.48}  \\
$\Delta$                                    & +0.07 & +0.06 & +0.47 & +0.20 & +0.10 & +0.30 & +0.17 & +0.23 & +0.23 & -0.03 & +0.53 & +0.23  \\
\bottomrule
\end{tabular}%
}
\label{tab:cross}
\end{table*}

\subsection{Cross-dataset evaluation}
As summarized in Table \ref{tab:cross}, DiscoVL achieves the competitive accuracy on ImageNet and shows superior performance in 6 out of the 10 datasets.
DiscoVL demonstrates superior performance, with improvements of 0.48\%, 0.48\% and 0.23\% for CoPrompt, Skip Tuning and MMRL, respectively.
This highlights that the performance gain primarily arises from our disentangled cross-modal representation aligner, which alleviates representation inconsistency across layers and enhances robustness to distribution shifts.

\subsection{Domain generalization}
We evaluate the robustness of different methods on out-of-distribution datasets by training models on the ImageNet source dataset and evaluating their performance on four ImageNet variants with domain shifts.
As shown in Table \ref{tab:domain}, our DiscoVL demonstrates significant performance across all scenarios.
DiscoVL achieves the best performance on ImageNet-A and improves the average target accuracy over PromptSRC by 0.18\%.
Furthermore, DiscoVL achieves the largest relative improvement of 0.47\% on ImageNet-R, which contains renditions with varied artistic styles, suggesting that DiscoVL effectively captures high-level semantic features that are invariant to style changes.

\begin{table}
\caption{Comparisons with state-of-the-art methods on domain generalization. On an average, DiscoVL achieves consistent improvement.}
\centering
\scalebox{0.7}{%
\begin{tabular}{@{}c|cccccc@{}}
\toprule
\multirow{2}{*}{\textbf{Method}} & \textbf{Source} & \multicolumn{5}{c}{\textbf{Target dataset}} \\
\cmidrule(r){2-7}
& ImageNet & ImageNetV2 & ImageNet-Sk & ImageNet-A & ImageNet-R & Average \\
\cmidrule(r){1-7}
CLIP \cite{radford2021learning}             & 66.73 & 60.83 & 46.15 & 47.77 & 73.96 & 57.18  \\
CoOp \cite{zhou2022learning}                & 71.51 & 64.20 & 47.99 & 49.71 & 75.21 & 59.28  \\
CoCoOp \cite{zhou2022conditional}           & 71.02 & 64.07 & 48.75 & 50.63 & 76.18 & 59.91  \\
MaPLe \cite{khattak2023maple}               & 70.72 & 64.07 & 49.15 & 50.90 & 76.98 & 60.27  \\
PromptSRC \cite{khattak2023self}            & 71.27 & 64.35 & \underline{49.55} & 50.90 & 77.80 & 60.65  \\
MMA \cite{yang2024mma}                      & 71.00 & 64.33 & 49.13 & 51.12 & 77.32 & 60.48  \\
AAPL \cite{kim2024aapl}                     & 71.37 & 64.20 & 48.80 & 50.60 & 76.87 & 60.12  \\
CoPrompt \cite{roy2024consistency}          & 70.80 & 64.25 & 49.43 & 50.50 & 77.51 & 60.42  \\
TextRefiner \cite{xie2025textrefiner}       & 72.06 & \underline{65.02} & 48.58 & 49.77 & 76.30 & 59.92  \\
SkipT. \cite{wu2025skip}                    & \textbf{72.77} & \textbf{65.67} & \textbf{49.73} & \underline{51.13} & \textbf{78.27} & \textbf{61.20}  \\
MMRL \cite{guo2025mmrl}                     & 72.03 & 64.47 & 49.17 & 51.20 & 77.53 & 60.59  \\
\midrule
\textbf{DiscoVL (Ours)}                       & \underline{72.10} & 64.60 & 49.40 & \textbf{51.30} & \underline{78.00} & \underline{60.83}  \\
$\Delta$                                    & +0.07 & +0.13 & +0.23 & +0.10 & +0.47 & +0.24  \\
\bottomrule
\end{tabular}%
}
\label{tab:domain}
\end{table}

\subsection{Few-shot learning}
\begin{figure*}[!t]
\centering
\includegraphics[scale=0.287]{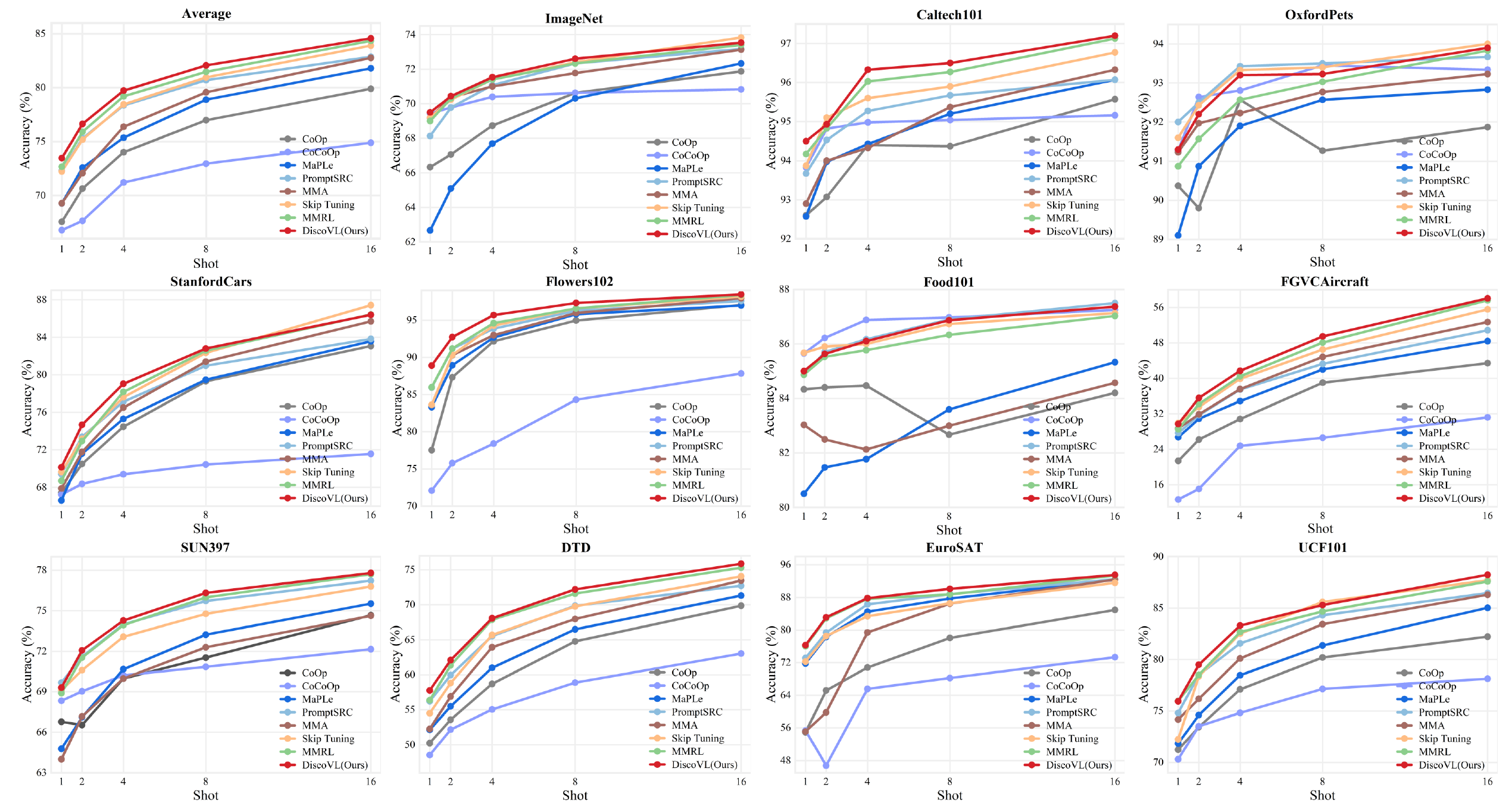}
\caption{Comparison with previous few-shot fine-tuning methods on 11 datasets under different shot settings. Our DiscoVL achieves new state-of-the-art performance.}
\label{fig.4}
\end{figure*}

This evaluation examines the model's transfer learning capability in limited-data scenarios.
Our model is trained on subsets of the training data with 1, 2, 4, 8, and 16-shots per class and subsequently evaluated on the full test sets.
As shown in Fig. \ref{fig.4}, under extreme few-shot conditions (1/2-shot), DiscoVL demonstrates strong advantages, achieving 73.48\% and 76.63\%, respectively, surpassing the previous best methods.
In the 16-shot setting, DiscoVL continues to achieve the highest performance at 84.58\%, exceeding PromptSRC and Skip Tuning by 1.71\% and 0.67\%, respectively.

\subsection{Ablation study}
\textbf{Core components.}
As shown in Table~\ref{tab:ablation1}, each component progressively contributes to performance improvement.
The multi-branch residual aligner and bidirectional feedback (DCRA) improve HM from 81.20\% to 81.32\%, while introducing orthogonal adversarial representation learning further increases HM to 81.58\%.
Both modules consistently improve performance when applied individually, and their combination yields the best generalization.

\textbf{Loss function.}
As shown in Table~\ref{tab:subtable1}, each loss component contributes incrementally: $L_{\text{atr}}^k$ improves HM by +0.20\%, adding $L_{\text{atr}}^{k'}$ yields a further +0.07\%, and incorporating orthogonal regularization $L_{\text{ortho}}$ achieves the best HM of 78.42\%, confirming that all terms are complementary.

\begin{table}[htbp]
  \centering
  \caption{Ablation study of DiscoVL components (average over 11 datasets) and loss function contributions (StanfordCars).}
  \begin{subtable}{0.48\textwidth}
    \centering
    \resizebox{\textwidth}{!}{
    \begin{tabular}{@{}c|ccc@{}}
    \toprule
    Method & Base & Novel & HM    \\
    \cmidrule(r){1-4}
    Baseline                      & 85.68 & 77.16 & 81.20  \\
    + Multi-branch res aligner    & 85.70 & 77.23 & 81.24   \\
    + Bidirectional feedback      & 85.77 & 77.30 & 81.32   \\
    + Adversarial representation  & 85.84 & 77.50 & 81.46    \\
    + Orthogonal regularization   & \textbf{85.98} & \textbf{77.61} & \textbf{81.58} \\
    \bottomrule
    \end{tabular}
    }
    \caption{Main components}
    \label{tab:ablation1}
  \end{subtable}
  \hfill
  \begin{subtable}{0.48\textwidth}
    \centering
    \resizebox{\textwidth}{!}{
        \begin{tabular}{@{}cccc|ccc@{}}
        \toprule
        Baseline &$L_{\text{atr}}^{k}$ &$L_{\text{atr}}^{k'}$ &$L_{\text{ortho}}^{k}{+}L_{\text{ortho}}^{k'}$ & Base & Novel & HM    \\
        \cmidrule(r){1-7}
        $\checkmark$ & $\times$ & $\times$ & $\times$             & 81.30 & 75.07 & 78.06 \\
        $\checkmark$ & $\checkmark$ & $\times$ & $\times$         & 81.57 & 75.20 & 78.26 \\
        $\checkmark$ & $\checkmark$ & $\checkmark$ & $\times$     & 81.70 & 75.23 & 78.33 \\
        $\checkmark$ & $\checkmark$ & $\checkmark$ & $\checkmark$ & \textbf{81.80} & \textbf{75.30} & \textbf{78.42} \\
        \bottomrule
        \end{tabular}
    }
        \caption{Loss function}
        \label{tab:subtable1}
  \end{subtable}
  \label{tab:ablation}
\end{table}

\begin{wrapfigure}[11]{r}{0.45\textwidth}
    \centering
    \begin{subfigure}[b]{0.2\textwidth}
        \centering
        \includegraphics[height=2cm]{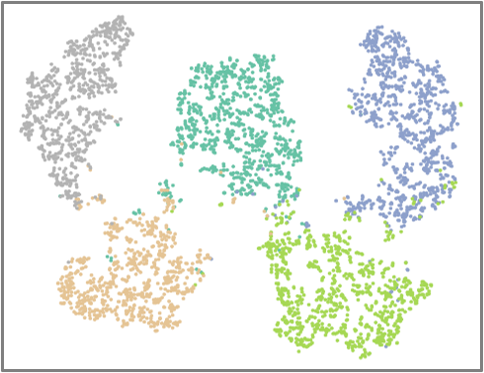}
        \caption{Baseline}
        \label{5-a}
    \end{subfigure}
    \hfill
    \begin{subfigure}[b]{0.2\textwidth}
        \centering
        \includegraphics[height=2cm]{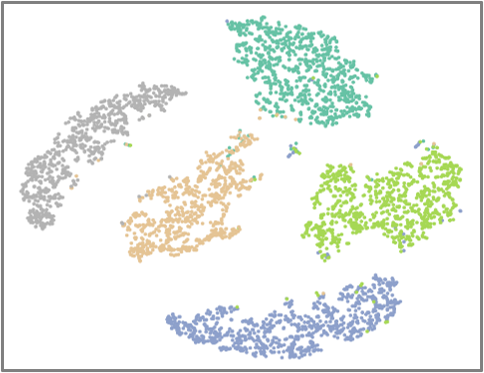}
        \caption{DiscoVL}
        \label{5-b}
    \end{subfigure}
    \caption{t-SNE visualization. Different colors denote different classes.}
    \label{fig.5}
\end{wrapfigure}
\subsection{Visualization}
\textbf{T-SNE}. As depicted in Fig.~\ref{fig.5}, we perform t-SNE visualizations on the EuroSAT to evaluate the effectiveness of DiscoVL.
The results show clear class-wise clustering, indicating that the learned feature representations are highly discriminative.
This is attributed to disentangled cross-modal representation aligner, which progressively integrates information from different modalities, mitigates inter-modal heterogeneity, and enhances the model's balanced understanding of semantic information.
Specifically, the disentanglement of shared and task-specific features ensures that common cross-modal knowledge is preserved while modality-unique characteristics are effectively isolated.
Furthermore, orthogonal and adversarial regularization reduce feature redundancy and enforce feature independence.

\begin{wrapfigure}[8]{r}{0.45\textwidth}
    \centering
    \includegraphics[width=0.45\textwidth]{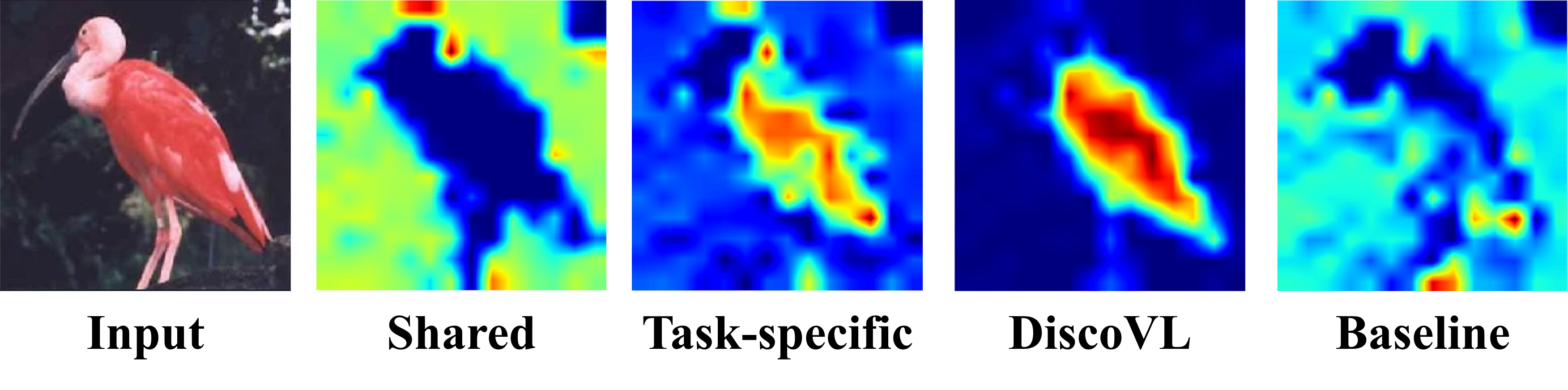}
    \caption{Grad-CAM visualization.}
    \label{fig.6}
\end{wrapfigure}
\textbf{Grad-CAM.}
In Fig. \ref{fig.6}, the shared branch yields broad, diffuse activations that capture class-agnostic contextual cues, while the task-specific branch produces more focused responses, pinpointing the most discriminative local regions for the target category.
Notably, our DiscoVL achieves the most refined and concentrated attention map.
This demonstrates its superior ability to synergistically integrate global context and local details, outperforming the baseline, whose attention remains comparatively scattered and less definitive.

\subsection{Training stability and efficiency}
We assess the training stability of DiscoVL by reporting standard deviations across three random seeds, as shown in Table \ref{tab:sta}.
DiscoVL demonstrates more stability than CoPrompt and MMRL in the base class on the ImageNet dataset.
In the novel class, DiscoVLs' stability is slightly lower than that of CoPrompt but shows improvement.
On the EuroSAT dataset, DiscoVL achieves the lowest standard deviation, indicating higher stability.
In the novel class, although the standard deviation increases compared to other methods, it still maintains relatively consistent performance.
The OARL and DCRA mechanisms introduced in DiscoVL effectively enhance training stability, leading to more reliable and consistent adaptation performance across different seed conditions.

\begin{table}
\centering
\caption{Training stability comparison, measured by mean $\pm$ standard deviation (\%). $\star$ means reproduced by ourselves.}
\label{training_stability}
\resizebox{0.7\textwidth}{!}{
    \begin{tabular}{@{}c|ccc|ccc@{}}
    \toprule
    \multirow{2}{*}{Method} & \multicolumn{3}{c|}{ImageNet} & \multicolumn{3}{c}{EuroSAT} \\
                            & Base     & Novel     & HM     & Base     & Novel    & HM    \\ \midrule
    Coprompt$^\star$           & $76.97 \pm 0.57$ & $71.10 \pm \textbf{0.00}$ & 73.92 & $93.67 \pm 1.08$ & $77.73 \pm 8.16$  & 84.96    \\
    MMRL                & $77.90 \pm 0.08$ & $71.30 \pm 0.28$ & 74.45 & $95.60 \pm 0.33$ & $80.17 \pm 5.05$  & 87.21   \\
    SkipT.$^\star$             & $77.77 \pm \textbf{0.09}$ & $70.40 \pm 0.22$ & 73.87 & $92.60 \pm 1.28$ & $82.10 \pm \textbf{2.91}$  & 87.03   \\
    DiscoVL (Ours)      & $78.07 \pm 0.13$ & $71.63 \pm 0.05$ & 74.71 & $96.80 \pm \textbf{0.21}$ & $81.20 \pm 2.99$  & 88.32    \\
    \bottomrule
    \end{tabular}
}
\label{tab:sta}
\end{table}

Table~\ref{tab:efficiency} compares the computational efficiency on ImageNet.
Overall, DiscoVL achieves a superior accuracy-efficiency trade-off compared to MMRL.
Specifically, it reduces the number of parameters by 55.7\%, lowers latency from 6.31 ms to 6.11 ms, improves throughput, and slightly decreases FLOPs. Despite its significantly reduced model complexity, DiscoVL still improves HM by 0.26\%, indicating that the proposed design enhances computational efficiency without sacrificing accuracy and even yields marginally better generalization.

\begin{table}[ht]
\centering
\caption{Efficiency on a single NVIDIA RTX 3060 GPU.}
\label{tab:efficiency}
\resizebox{0.7\textwidth}{!}{
\begin{tabular}{llllllll}
\toprule
Method & Param. (M) & Lat.(ms) & FPS & FLOPs (G) & Train time (min) & HM (\%) \\
\midrule
MMRL           & 4.99               &  6.31             & 158.44              & 21.58              & 156 & 74.45         \\
DiscoVL (Ours) & 2.21$_{(-2.78)}$   &  6.11$_{(-0.20)}$  & 163.68$_{(+5.24)}$  & 21.50$_{(-0.08)}$  & 147$_{(-9)}$ & 74.71$_{(+0.26)}$   \\
\bottomrule
\end{tabular}
}
\end{table}

\section{Conclusion}
\label{Conclusion}
We identify representation entanglement, the conflation of class-specific and class-agnostic information, as a principal bottleneck of existing VLMs prompt-learning methods and propose DiscoVL, a disentangled cross-modal representation learning framework that explicitly factorizes the adapted feature space into shared and task-specific subspaces.
DCRA separates and progressively aligns cross-modal cues through multi-branch residual structure and bidirectional feedback, while OARL preserves the class-agnostic representation via orthogonal adversarial constraints that suppress token collapse.
Across 15 benchmarks, we observe consistent performance gains on both base and novel classes. Dedicated disentanglement analyses further confirm that the separated components are genuinely distinct and functionally complementary.
Future work will explore coupling DCRA with dynamic prompt schedulers and extending orthogonal adversarial learning to sequential or streaming settings.

\section*{Acknowledgements}
This work was supported by the Key R \& D Program of Shandong Province, China under Grant No.2025CXPT094, 20 Guidelines for New Colleges in Jinan City under Grant No. 202333044, Qilu University of Technology (Shandong Academy of Sciences) Major Innovation Project of Science, Education and Industry Integration Pilot Project 2025ZDZX12, and the Natural Science Foundation of Shandong Province Young Scientists Fund under Grant No. ZR2026QC1572.
\bibliographystyle{splncs04}
\bibliography{main}

\clearpage
\appendix
\begin{center}
{\Large\bfseries Supplementary Material}
\end{center}
\setcounter{section}{0}
\setcounter{table}{0}
\setcounter{figure}{0}
\renewcommand{\thesection}{S\arabic{section}}
\renewcommand{\thetable}{S\arabic{table}}
\renewcommand{\thefigure}{S\arabic{figure}}

\section{Hyperparameter Analysis}
\label{sec:supp_hyperparam}
\textbf{Effectiveness of value $\gamma$, $\alpha$ and $\beta$.}
We perform an ablation study on the balancing parameter $\gamma$ in the orthogonal adversarial regularization.
From the results in Table~\ref{tab:supp_gamma}, the optimal $\gamma$ is 0.3, while values of 0.2 and 0.4 result in slight performance degradation or no further improvement, respectively, indicating that overly strong regularization yields no additional benefits.
The parameter $\alpha$ regulates the learning process by balancing the contributions of representation token features and class token features.
As presented in Table~\ref{tab:supp_alpha}, performance generally improves with increasing $\alpha$, with the HM peaking at 84.46\% when $\alpha$ is 0.7.
The penalty coefficient $\beta$ controls the strength of regularization, promoting alignment between class token features and CLIP's fixed feature space.
As depicted in Table~\ref{tab:supp_beta}, raising $\beta$ from 0.0 to 0.5 slightly improves both base and novel class accuracy, yielding a peak HM of 41.60.
Further increasing $\beta$ to 2.0 results in a marginal performance drop.

\vspace{-0.8cm}
\begin{table}[ht]
\centering
\caption{Ablation studies on $\gamma$, $\alpha$ and $\beta$.}
\vspace{-0.2cm}
\label{tab:supp_ablation}
\subfloat[Food101]{
\label{tab:supp_gamma}
\scalebox{0.7}{%
\begin{tabular}{@{}c|ccc@{}}
\toprule
$\gamma$ & Base & Novel & HM \\
\midrule
0.1 & 90.67 & 91.60 & 91.13 \\
0.2 & 90.40 & 91.57 & 90.98 \\
\textbf{0.3} & \textbf{90.73} & \textbf{91.67} & \textbf{91.20} \\
0.4 & 90.50 & 91.60 & 91.10 \\
\bottomrule
\end{tabular}%
}
}
\quad
\subfloat[UCF101]{
\label{tab:supp_alpha}
\scalebox{0.7}{%
\begin{tabular}{@{}c|ccc@{}}
\toprule
$\alpha$ & Base & Novel & HM \\
\midrule
0.0 & 88.53 & 80.47 & 84.31 \\
0.3 & 88.40 & 80.23 & 84.12 \\
0.5 & 88.60 & 80.50 & 84.36 \\
\textbf{0.7} & \textbf{88.67} & \textbf{80.63} & \textbf{84.46} \\
\bottomrule
\end{tabular}%
}
}
\quad
\subfloat[FGVCAircraft]{
\label{tab:supp_beta}
\scalebox{0.7}{%
\begin{tabular}{@{}c|ccc@{}}
\toprule
$\beta$ & Base & Novel & HM \\
\midrule
0.0 & 46.37 & 37.20 & 41.29 \\
0.2 & 46.40 & 37.53 & 41.50 \\
\textbf{0.5} & \textbf{46.40} & \textbf{37.70} & \textbf{41.60} \\
2.0 & 46.37 & 37.60 & 41.53 \\
\bottomrule
\end{tabular}%
}
}
\end{table}

\vspace{-0.5cm}
\textbf{Regularization strategies.}
We explore regularization strategies to align class tokens with frozen CLIP features on EuroSAT.
As summarized in Table~\ref{tab:supp_reg}, cosine regularization achieves the best performance with an HM of 88.32\%.
$L_1$ regularization attains the lowest HM, while MSE regularization provides a competitive HM.

\textbf{Effect of branch $E$}.
Table \ref{tab:subspace_E} on EuroSAT shows $E=4$ is optimal, and $E=2$ underperforms across all metrics, and $E=6$ yields no further gain while adding cost.
We adopt $E=4$ as the default.

\vspace{-0.5cm}
\begin{table}[htbp]
\setlength{\abovecaptionskip}{1pt}
\setlength{\belowcaptionskip}{1pt}
\centering
\begin{minipage}[t]{0.4\linewidth}
\centering
\caption{Regularization}
\footnotesize
\setlength{\tabcolsep}{1pt}
\label{tab:supp_reg}
\scalebox{0.7}{%
\begin{tabular}{@{}c|ccc@{}}
\toprule
Regularization & Base & Novel & HM \\
\midrule
\textbf{Cosine} & \textbf{96.80} & \textbf{81.20} & \textbf{88.32} \\
$L_1$ & 96.50 & 80.87 & 88.00 \\
MSE & 96.63 & 81.00 & 88.13 \\
\bottomrule
\end{tabular}
}
\end{minipage}%
\hspace{0.05\linewidth}%
\begin{minipage}[t]{0.4\linewidth}
\centering
\caption{Value of $E$}
\footnotesize
\setlength{\tabcolsep}{1pt}
\label{tab:subspace_E}
\scalebox{0.7}{%
\begin{tabular}{ccc c}
\toprule
$E$ & Base & Novel & HM \\
\midrule
2   & 96.61 & 81.07 & 88.17 \\
\textbf{4}   & \textbf{96.80} & \textbf{81.20} & \textbf{88.32} \\
6   & 96.47 & 80.93 & 88.02 \\
\bottomrule
\end{tabular}
}
\end{minipage}
\end{table}

\vspace{-0.3cm}
\textbf{Gate behavior analysis.}
As shown in Table~\ref{tab:gate}, the fusion weights remain stable between 0.723 and 0.731 across all layers, indicating a strong cross-modal fusion strategy, where approximately 73\% of the information is derived from the other modality.
This consistently high gating value demonstrates effective bidirectional information flow and stable fusion strength between the text and visual modalities across different network depths.
\vspace{-0.5cm}
\begin{table}[ht]
\centering
\caption{Fusion weights analysis.}
\scalebox{0.7}{%
\label{tab:gate}
\resizebox{0.9\linewidth}{!}{
\begin{tabular}{lllllllll}
\toprule
Layer index & 1 & 2 & 3 & 4 & 5 & 6 & 7 & mean\\
\midrule
Fusion weights & 0.7276  & 0.7273  & 0.7228  & 0.7294 & 0.7276 & 0.7276  & 0.7306  & 0.7276     \\
\bottomrule
\end{tabular}
}
}
\end{table}

\section{Disentanglement Analysis}
\label{sec:supp_disentangle}
\label{sec:disentangle_summary}

\textbf{Separability.}
As shown in Table~\ref{tab:subspace_sep}, the per-layer mean absolute cosine similarity between the shared and task-specific branches remains below 0.10 across all layers, demonstrating that their features are highly orthogonal with minimal overlap.
The task-specific branch consistently outperforms the shared branch in linear probe accuracy, indicating that it encodes class-discriminative information, whereas the shared branch preserves class-agnostic representations.
The stability of this trend across layers substantiates that DCRA achieves robust functional disentanglement at varying network depths.

\begin{table}[htbp]
\centering
\caption{Separability across different layers (averaged over 11 datasets). 'Cos.\ Sim.' is the mean absolute cosine similarity between shared and task-specific branch features. 'Probe Acc.' is the base-class accuracy of a frozen linear probe trained on each branch independently.}
\label{tab:subspace_sep}
\scalebox{0.7}{%
\begin{tabular}{@{}c|c|cc@{}}
\toprule
Layer & Cos.\ Sim.\ ($\downarrow$) & Shared Probe Acc.\ (\%) & Task-Specific Probe Acc.\ (\%) \\
\midrule
6  & 0.08 & 67.42 & 81.35 \\
7  & 0.07 & 68.11 & 82.04 \\
8  & 0.09 & 67.89 & 82.47 \\
9  & 0.06 & 68.53 & 82.71 \\
10 & 0.08 & 68.20 & 82.13 \\
11 & 0.07 & 68.74 & 81.96 \\
\bottomrule
\end{tabular}
}
\end{table}

\textbf{Branch intervention.}
Table~\ref{tab:branch_mask} provides the branch intervention results.
Zeroing-out the task-specific branch drops base-class accuracy by 6.2\% but novel by only 1.2\% in ImageNet; masking the shared branch barely affects base ($-$0.9\%) but hurts novel generalization ($-$4.4\%), validating functional disentanglement.
This trend holds consistently across all three datasets, validating the functional disentanglement of the two branches.

\begin{table}[htbp]
\centering
\caption{Branch intervention results. 'Full' denotes the proposed DiscoVL. '$\backslash$Task' and '$\backslash$Shared' denote the task-specific and shared branches, respectively.}
\label{tab:branch_mask}
\scalebox{0.7}{%
\begin{tabular}{@{}l|ccc|ccc|ccc@{}}
\toprule
& \multicolumn{3}{c|}{ImageNet} & \multicolumn{3}{c|}{EuroSAT} & \multicolumn{3}{c}{UCF101} \\
Setting & Base & Novel & HM & Base & Novel & HM & Base & Novel & HM \\
\midrule
Full           & 78.07 & 71.63 & 74.71 & 96.80 & 81.20 & 88.32 & 88.67 & 80.63 & 84.46 \\
$\backslash$Task   & 71.90 & 70.47 & 71.18 & 89.53 & 79.80 & 84.39 & 81.20 & 79.43 & 80.31 \\
$\backslash$Shared & 77.13 & 67.20 & 71.82 & 96.07 & 76.53 & 85.19 & 87.93 & 76.10 & 81.59 \\
\bottomrule
\end{tabular}
}
\end{table}

\end{document}